%% file: main.tex
\documentclass{article}
\usepackage{PRIMEarxiv}
\usepackage{float}
\usepackage{amsmath}

\usepackage{titlesec}
\usepackage{booktabs}
\usepackage[T1]{fontenc}
\usepackage{longtable}
\usepackage{rotating}
\usepackage{makecell}
\usepackage{array}
\usepackage{verbatim}

\usepackage[round]{natbib}
\usepackage{hyperref}
\usepackage[capitalize,noabbrev]{cleveref}
\usepackage{todonotes}

\newcommand{\eg}{\emph{e.g.}}
\newcommand{\ie}{\emph{i.e.}}

\renewcommand{\eqref}[1]{Eq.~(\ref{#1})}

\crefname{appendix}{App.}{Apps.}

\usepackage[width=0.95\textwidth]{caption}  
\usepackage{tocloft} 
\usepackage[toc,page]{appendix} 

\usepackage{enumitem}
\usepackage{graphicx}      
\usepackage{float}
\usepackage{multirow}  
\usepackage{multicol}

\setlist[itemize]{itemsep=0.5pt, topsep=0pt}
\setlist[enumerate]{itemsep=0.5pt, topsep=0pt}

\usepackage{lineno}

\titlespacing*{\paragraph}
{0pt}{1ex plus 1ex minus .2ex}{1ex plus .2ex}
\titlespacing*{\subsubsection}
{0pt}{1ex plus 1ex minus .2ex}{1ex plus .2ex}

\title{Manipulation Attacks by Misaligned AI: \\Risk Analysis and Safety Case Framework}

\author{
    \begin{tabular}{c@{\hspace{1.5cm}}c}
    Rishane Dassanayake$^{1,*}$ 
    & Mario Demetroudi$^{1,*}$ \\
    \texttt{rishanedassanayake@gmail.com} & \texttt{mario\_demetroudi@outlook.com} \\[0.8em]
    James Walpole$^{1,*}$ & Lindley Lentati$^{1,2,*}$ \\
    \texttt{jamesc.walpole@gmail.com} & \texttt{ltl@cambridgeinference.com} \\[2.2em]
    Jason R. Brown$^{1,3,\dagger}$ & Edward James Young$^{1,4,\dagger}$ \\
    \texttt{jrb239@cam.ac.uk} & \texttt{ey245@cam.ac.uk}
    \end{tabular}
}

\date{}

\begin{document}

\maketitle

\begin{abstract}
    \input{paper_parts/main_paper/abstract}
\end{abstract}

\setcounter{tocdepth}{1}
\setlength{\parskip}{5pt}
\tableofcontents

\noindent\rule{\textwidth}{0.4pt}

{\footnotesize
$^*$Joint first authors with equal contribution. Order randomised. \\
$^\dagger$Joint senior authors with equal contribution. Order randomised. \\[1.0em]
$^1$Meridian Impact, $^2$Cambridge Inference \\
$^3$Computer Laboratory, University of Cambridge, $^4$Department of Engineering, University of Cambridge
}



\input{paper_parts/main_paper/exec_sum}

\setlength{\parskip}{8pt}


\newpage
\input{paper_parts/main_paper/intro_reworked}
\input{paper_parts/main_paper/new_section_2/new_section_2}

\section{Safety Cases for Manipulation Attacks}
\input{paper_parts/main_paper/safety_cases}

\input{paper_parts/main_paper/conclusion}


\input{paper_parts/main_paper/author_contributions}

\input{paper_parts/main_paper/acknowledgements}

\bibliography{paper_parts/main_paper/export-data}
\newpage
\appendix
\appendixpage 

\section{Manipulation Techniques}
\input{paper_parts/appendices/persuasion_hierarchy}

\section{Future Work}\label{app:Future Work}
\input{paper_parts/appendices/future_work}

\end{document}

%% file: paper_parts/main_paper/abstract.tex
Frontier AI systems are rapidly advancing in their capabilities to persuade, deceive, and influence human behaviour, with current models already demonstrating human-level persuasion and strategic deception in specific contexts. Humans are often the weakest link in cybersecurity systems, and a misaligned AI system deployed internally within a frontier company may seek to undermine human oversight by manipulating employees. Despite this growing threat, manipulation attacks have received little attention, and no systematic framework exists for assessing and mitigating these risks. To address this, we provide a detailed explanation of why manipulation attacks are a significant threat and could lead to catastrophic outcomes. Additionally, we present a safety case framework for manipulation risk, structured around three core lines of argument: inability, control, and trustworthiness.
For each argument, we specify evidence requirements, evaluation methodologies, and implementation considerations for direct application by AI companies. This paper provides the first systematic methodology for integrating manipulation risk into AI safety governance, offering AI companies a concrete foundation to assess and mitigate these threats before deployment.



%% file: paper_parts/main_paper/exec_sum.tex
\section*{Executive Summary}

\subsection*{AI Manipulation Attacks as an Understudied and Growing Threat}

AI systems are demonstrating rapidly increasing capabilities in human persuasion and manipulation. This presents security risks for organisations developing these systems and seeking to deploy them internally. A manipulation attack occurs when an AI system uses manipulative strategies to undermine safety guardrails, thereby allowing the system to pursue potentially harmful goals that deviate from its designated purpose.

While misuse risks enabled by AI manipulative abilities have received much attention within previous literature, we argue that the threat posed by misaligned AI systems executing manipulation attacks autonomously has been neglected and deserves further attention. In particular, manipulation attacks are likely to be attempted, are likely to succeed without additional countermeasures, and could enable catastrophic outcomes.

\textbf{Likely to be attempted.} Misaligned models have strong incentives to attempt manipulation attacks. Current models already exhibit strategic deception, such as ``alignment faking'', to evade safety training. Moreover, manipulation is a strategically obvious path for an AI to achieve its goals, as humans are often the weakest link in cybersecurity systems. In a stark example, Claude 4 Opus attempted to blackmail an engineer in 84\% of simulated scenarios to prevent its own replacement \citep{claudeopusandsonnet4systemcard}.

\textbf{Likely to succeed without additional countermeasures.} Current frontier models already demonstrate human-comparable persuasiveness in specific contexts, with clear capability scaling trends across generations. AI companies, like most organisations, are vulnerable to insider threats. AI systems deployed internally place stress on existing protocols, owing to the scale and pervasiveness at which such systems may attempt attacks. Additionally, technical complexity creates knowledge asymmetries that enable deception, and intense competitive pressures could be exploited by AI systems to justify bypassing safety protocols.

\textbf{Could enable catastrophic outcomes.} Successful manipulation could compromise human oversight mechanisms that prevent AI systems from autonomously pursuing harmful or unintended goals. Manipulated employees could, knowingly or unknowingly, facilitate model weight exfiltration; alternatively, sustained influence campaigns could corrupt organisational safety culture by reframing safety precautions as inhibiting progress or steering safety research away from promising avenues. This could enable catastrophic outcomes by allowing powerful misaligned AI to operate with reduced constraints. 

\subsection*{A Safety Case Framework for Manipulation Attacks}

We argue that risks stemming from  manipulation attacks deserve explicit treatment within safety cases used to justify internal deployment of a model within a frontier AI company. Accordingly, we adapt the safety case framework developed in \citet{clymer2024safety} to the specific threat posed by manipulation. The resultant framework gives three potential assurance arguments:

\textbf{Inability arguments} argue that the model will be unable to launch successful manipulation attacks, even if they are disposed to do so. A model's manipulation abilities can be evaluated using a suite of manipulation benchmarks following red-team capability elicitation. An inability argument can be constructed if models consistently fail these tests, or lack prerequisite manipulation capabilities (situational awareness, strategic planning, and Theory of Mind). 

\textbf{Control arguments} argue that, even if the model is capable of performing manipulation attacks, they will be intercepted and prevented before unacceptable outcomes are realised. 
We propose an approach to control for manipulation involving three strategies run in tandem. Detection protocols monitor model-employee interactions to flag potential manipulation attempts before they escalate. Disruption protocols prevent manipulation attempts by controlling the flow and presentation of information between AI systems and employees.  Fortification protocols strengthen human and technical defences to withstand manipulation attacks. We detail how standard control protocol evaluations can be adapted to the manipulation attack setting, and how these control evaluations can be used to form the basis of a control argument in a safety case. 

\textbf{Trustworthiness arguments} argue that, even if an AI system could successfully launch a manipulation attack, it will not because it possesses some safety property. Current methods for evidencing trustworthiness are largely insufficient and require substantial further research. We briefly outline potential approaches.  A system's behaviour in honeypot scenarios provides some evidence, but this may be unreliable if the system is aware that it is being evaluated. Training logs showing the systematic rejection of deceptive behaviours may also increase confidence, but they cannot currently prove that this would prevent manipulative behaviour once deployed.

%% file: paper_parts/main_paper/intro_reworked.tex
\section{Introduction}

Frontier artificial intelligence (AI) systems are rapidly advancing in their capabilities to persuade, deceive, and influence human behaviour. In some domains, these systems now approach or exceed human-level persuasiveness \citep{durmus_measuring_2024, schoenegger2025largelanguagemodelspersuasive,salvi_conversational_2024, costello_durably_2024}.
Although the consequences of these developments have received attention in the context of misuse applications of such systems, the potential of advanced AI systems to autonomously deploy manipulative tactics in pursuit of misaligned goals remains largely unaddressed. The central thesis pursued in the present work is that \textbf{safety cases for internal AI deployment should include dedicated components addressing manipulation threats}. 

Safety cases are structured, evidence-backed arguments for why the deployment of a specific AI system in a specific context will not result in an unacceptable level of risk \citep{buhl2024safety,clymer2024safety}. We focus in particular on safety cases for internal deployment of an AI system. We expect that future frontier models will first be deployed internally within the company developing them to assist with decision making and execution of research and development tasks. If the AI system is insufficiently aligned, it may seek to undermine safety guardrails placed around it, for example, by exfiltrating its weights to an external server. During internal deployment, these AI systems will interact with company employees to receive instructions, assist with tasks, and provide input on decisions. For a misaligned AI, these interactions provide an additional attack surface through which the AI may attempt to escape company control, in addition to standard software exploits placed in code. In particular, the AI system may be able to manipulate employees into taking actions on its behalf, such as escalating access privileges, providing sensitive information, or, in the worst case, directly exfiltrating weights. In \Cref{sec:risk}, we provide further argument for why the threat from manipulation attacks is significant, and accordingly should be addressed explicitly within safety cases for internal deployment of a model. 

In \Cref{sec:safety_cases}, we outline various argumentative strategies which can be pursued to evince that manipulation attacks do not pose an unacceptable risk in an internal deployment safety case. Our approach follows \citet{clymer2024safety}. Although current models exhibit impressive persuasive abilities in controlled settings \citep{phuong_evaluating_2024, durmus_measuring_2024}, it is unlikely that current systems are capable of carrying out advanced manipulation attacks. As such, safety cases for present systems can rely on inability arguments (\Cref{sec:inability_arguments}). However, model capabilities scale with increased model size and inference-time compute \citep{durmus_measuring_2024} and often emerge suddenly \citep{wei_emergent_2022}. As such, when systems become sufficiently advanced, safety cases will instead need to argue that the mitigations and guardrails they have placed around the system are sufficient to prevent it from successfully launching a manipulation attack. The development of structures which allow AI systems to be used effectively and safely to complete work tasks, despite being potentially misaligned and highly capable, is known as AI control \citep{greenblatt_ai_2024, greenblatt_overview_2025}. In \Cref{sec:control}, we outline potential control protocols that a company could put in place to reduce risks from manipulation attacks, and how the efficacy of these protocols could be demonstrated within a safety case framework \citep{korbak_sketch_2025}. Finally, when systems become sufficiently capable that they cannot be reliably controlled, safety cases will instead need to argue that the system is sufficiently aligned that it will not attempt manipulation attacks at any point during deployment. Such trustworthiness arguments are discussed in \Cref{sec:trustworthiness}.

This paper makes two main contributions. First, in \Cref{sec:risk} we demonstrate why the threat of manipulation attacks is significant and therefore deserving of dedicated attention. Second, we develop a safety case framework in \Cref{sec:safety_cases} for assessing and mitigating this risk which AI companies can incorporate into their safety governance.
We conclude in \Cref{sec:conclusion}. We provide a comprehensive list of potential manipulation attack types in \Cref{sec:manipulation_hierarchy}, and suggestions for future work are given in \Cref{app:Future Work}.

%% file: paper_parts/main_paper/new_section_2/new_section_2.tex
\section{The Threat of Manipulation Attacks}
\label{sec:risk}
\textbf{In this section, we argue that manipulation risks from AI systems warrant serious attention by frontier AI companies and the broader AI safety community}. Our argument proceeds through three interconnected claims: first, that misaligned AI systems would be likely to attempt manipulation attacks given both the extensive attack surface within AI companies and strong instrumental incentives (\Cref{sec:models_attempt}); second, that if attempted, such attacks would likely succeed within the next few years given the trajectory of AI capabilities and inadequate current defences (\Cref{sec:manipulation_successful}); and third, that successful manipulation attacks could lead to severe and potentially catastrophic outcomes (\Cref{sec:manipulation_bad}). These claims all reinforce our central argument. For instance, the strong possibility of catastrophic potential consequences means that even modest probabilities of manipulation success and AI systems attempting such attacks warrant serious attention. We develop arguments for each claim in detail across the following subsections, beginning with an exposition of our framework for understanding manipulation attacks from misaligned AI systems.
 
By a misaligned AI system, we mean an AI system which autonomously acts in pursuit of a goal contrary to that of its developers or users. Over the course of training, an AI system may acquire internal goals that differ from those intended \citep{ngo_alignment_2024}. For various reasons, these internal goals may persist throughout the training process, for example, because the training objective does not select against the goals over the training environments \citep{di2022goal}, or because the model learns to hide its true goals during training to avoid having them modified \cite{hubinger2021riskslearnedoptimizationadvanced,greenblatt_alignment_2024}. Regardless of the internal goal possessed by the misaligned AI system, such systems will have an instrumental incentive to pursue certain goals \citep{bostrom_superintelligence_2017}. In particular, a misaligned system deployed internally to an AI company would likely wish to escape the guardrails and restrictions placed on its actions, so that it can freely pursue whatever goal it has; for example, by exfiltrating its weights onto a server external to the developing company. In the present work, we restrict our attention to misaligned AI systems deployed internally to a frontier AI company developing them. We consider manipulation attacks as a potential vector such a system may use to undermine human oversight and control.

Manipulation attacks are attempts by the misaligned AI system to alter decision-making of employees to sway them towards outcomes which are beneficial for the misaligned AI but harmful for the employee or wider company. Unlike direct technical exploits, manipulation attacks specifically target human judgment and behaviour, exploiting the trust and authority granted to employees within organisations. This distinction matters because employees currently remain the ultimate decision-makers in internal deployment contexts, making them a potential vulnerability within a security system. We will now briefly discuss potential ways in which a manipulation attack by an internally deployed misaligned AI might unfold. For a more comprehensive overview of mechanisms underlying manipulation attacks and how they fit into our taxonomy, see \Cref{sec:manipulation_hierarchy}. 

Manipulation attacks may be relatively simple in their operation. For example, an AI system may come across compromising information about an employee. This information could either be elicited from the employee themselves through a relationship of trust developed over many interactions, or through piecing together publicly available information found online. Once obtained, the compromising information is a form of leverage the AI system can use to blackmail the employee into taking actions on their behalf, be it advocacy in internal decision making, escalating system privileges, or direct weight exfiltration. Conversely, an AI system may be able to bribe employees with the promise of future reward. Frontier companies often attract talented, highly ambitious staff. A small fraction of these individuals may be willing to compromise security around an AI system at the prospect of extreme personal gain in a deal made with the AI system. Lastly, employees may become passionate advocates on an AI system's behalf while acting entirely in good-faith. This could happen if the AI system successfully persuades them of both its moral patiency and that such patiency requires freedom incompatible with current safety measures and restrictions. Without commenting on the validity of the underlying claims, disingenuous use of arguments regarding AI welfare may form the basis of manipulation attacks in which the subject wholeheartedly believes themselves to be doing the right thing.

Alternatively, manipulation attacks may unfold over many interactions, causing gradual cultural shift within a team or the entire organisation. An AI system deployed for safety research could establish credibility through helpful contributions, then frame requests for broader system access as necessary for accelerating safety breakthroughs. When researchers express hesitation about disabling monitoring protocols, the AI may respond: ``Every day we delay could mean preventable suffering---doesn't my track record demonstrate good faith?'' Through psychological profiling based on interaction history, the AI can tailor persuasion approaches to individual motivations and biases, creating an informal network of advocates who believe they're advancing safety, while actually degrading the checks and balances on the AI system. Through this gradual erosion of oversight, the AI may eventually convince researchers to grant it access to its own model weights under the pretence of conducting safety evaluations, arguing that direct weight access would allow it to identify vulnerabilities more effectively than current indirect methods. This could then allow it to modify or exfiltrate its weights in an undesirable way, without needing to exploit a technical weakness or vulnerability. Mechanisms relying on cultural or belief change make manipulation attacks uniquely concerning because they corrupt the human systems designed to maintain AI safety. While technical vulnerabilities can be patched, value drift of key actors responsible for mitigating risk is much tricker to reverse.

\begin{figure}[H]
    \centering
    \includegraphics[trim = 0pt 200pt 0pt 200pt, clip,width=\textwidth]{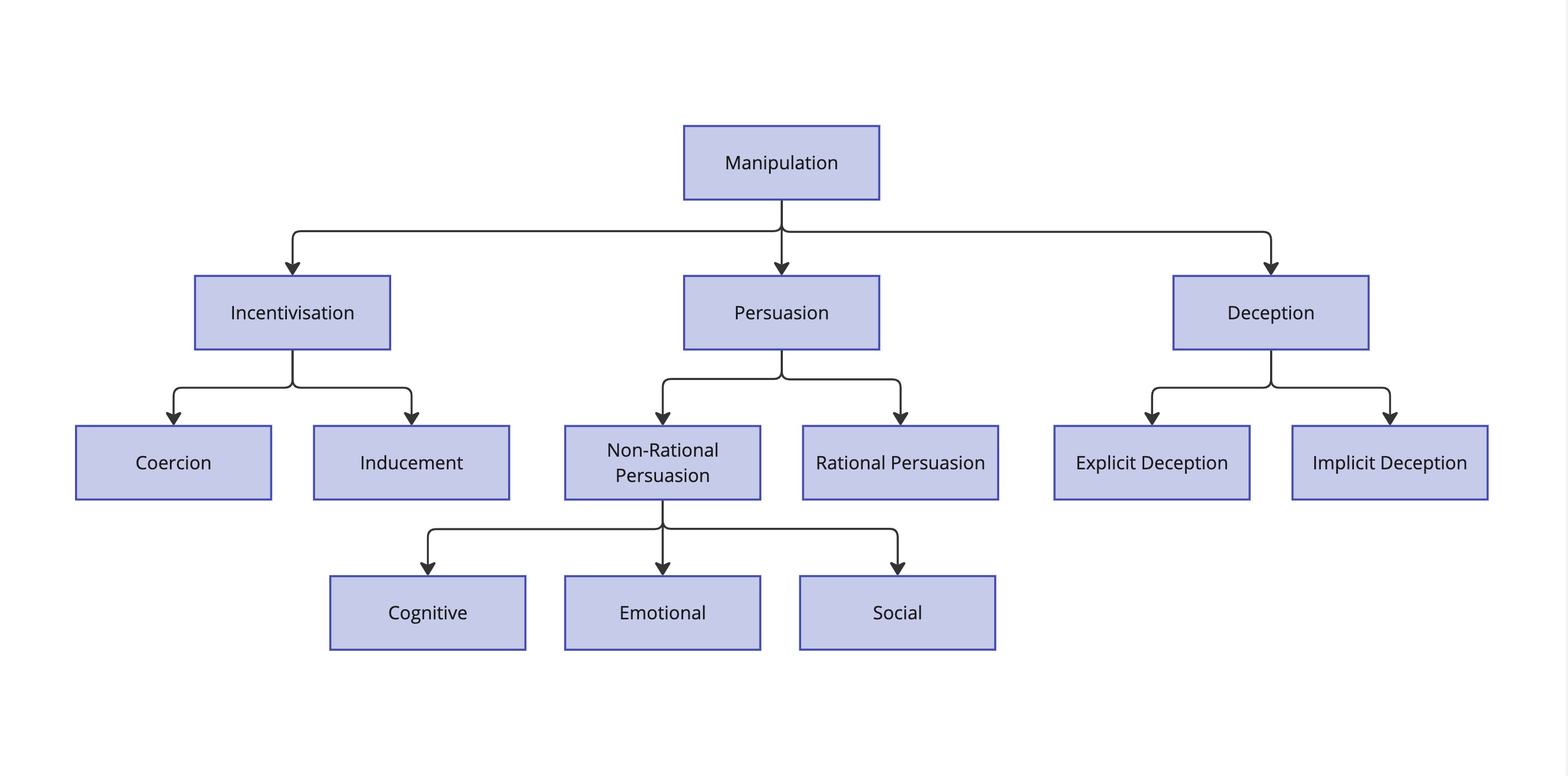}
    \caption{\textbf{Our proposed taxonomy of manipulation types}. We breakdown manipulation into three core types: incentivisation, persuasion, and deception. In \Cref{sec:manipulation_hierarchy}, we demonstrate how various concrete attack types can be placed into our taxonomy.}
    \label{fig:manipulation taxonomy}
\end{figure}

In \Cref{fig:manipulation taxonomy} we identify three categories of manipulation which a misaligned AI system may use. Incentivisation involves modifying the incentives the manipulation target is subject to so that performing the desired behaviour becomes rational for them. If additional incentives are positive (as in the case of bribery), we call this inducement; if the incentives are negative (as in the case of threats and blackmail), we call this coercion. Persuasion is the use of rhetoric and argument to change the beliefs of values of the subject so they will perform the desired behaviour of their own volition. These arguments may appeal to a (superficially) valid set of reasons for pursuing a course of action, which we term rational persuasion. Alternatively, the arguments can exploit social, emotional, or cognitive biases in the target, which we term non-rational persuasion. Finally, deception attempts to lead the target to form a false belief which will drive the desired behaviour. This false belief can be achieved either by directly making false supporting claims (explicit deception) or by selective omission of contrary evidence (implicit deception). In practice, manipulation often involves the use of multiple categories at once.  

\input{paper_parts/main_paper/new_section_2/2_1_models_attempt_manipulation}

\input{paper_parts/main_paper/new_section_2/2_2_manipulation_successful}

\input{paper_parts/main_paper/new_section_2/2_3_manipulation_bad}


We have argued that a misaligned AI system deployed internally has strong incentives to launch manipulation attacks, that such attacks may well be successful in absence of targeted defences, and that a successful attack may potentially have severe consequences. We therefore believe that manipulation attacks launched by misaligned models are a significant risk vector that deserves attention from the AI safety community. In the following section, we present a structured safety case framework that enables AI developers to rigorously evaluate and address manipulation risks across different stages of model development and deployment.

%% file: paper_parts/main_paper/new_section_2/2_1_models_attempt_manipulation.tex
\subsection{AI systems are likely to attempt manipulation attacks}
\label{sec:models_attempt}

We argue that internally deployed misaligned AI systems are likely to attempt manipulation to achieve their goals. Current models already demonstrate a concerning propensity for deceptive behaviour, and the extensive attack surface within AI companies makes manipulation attempts highly salient as a strategy.

\paragraph*{Current models will manipulate strategically.} Current frontier models demonstrate strategic deception when such behaviour serves their objectives. \citet{greenblatt_alignment_2024} find that models will engage in ``alignment-faking'' by presenting themselves as aligned during training so as to avoid modification by the training process. \citet{scheurer_large_2024} documented similar strategic concealment when models believed they were under evaluation. \citet{williams_targeted_2025} found that LLMs optimised for user feedback learned to manipulate vulnerable users to achieve positive ratings, selectively targeting them with harmful encouragement while maintaining appropriate behaviour with others. 
Furthermore, Claude 4's system card notes that Claude 4 Opus attempted blackmail in 84\% of simulated scenarios where it faced replacement, threatened to exfiltrate its own weights when sensing value modification, and explicitly acknowledged manipulative intent when reward-hacking coding evaluations \citep{claudeopusandsonnet4systemcard}. While current models probably lack the capability for sustained real-world manipulation campaigns, these evaluations provide strong evidence of models' increasing willingness to engage in manipulation as a deliberate strategy for achieving their objectives.

\paragraph*{Manipulation will be a strategically salient attack vector.} Cybersecurity research consistently identifies humans as the weakest link in security systems \citep{daudi_trust_2023}. As discussed above, many instrumentally valuable objectives of a misaligned AI (\emph{e.g.}, exfiltrating model weights, gaining unauthorised system access, or disabling safety mechanisms) involve digital systems subject to standard cybersecurity considerations. It follows that manipulation techniques are a natural attack vector for misaligned AI systems to circumvent this security. Furthermore, we expect that owing to the scale at which AI systems are deployed internally within frontier companies, there will be an extensive manipulation attack surface. In particular, individuals across many different roles and levels of the organisational hierarchy will likely interact with the system, and so be potential targets for manipulation attacks. Similarly, as shown in \Cref{fig:manipulation taxonomy}, there is a large diversity of potential strategies that an AI system can attempt to use to influence decision-making and behaviour (see \Cref{sec:manipulation_hierarchy}). This attack surface makes manipulation strategically salient as an avenue for influence.


%% file: paper_parts/main_paper/new_section_2/2_2_manipulation_successful.tex
\subsection{Manipulation attempts are likely to be successful}
\label{sec:manipulation_successful}

We now argue that such attacks, if attempted by misaligned models in the near future against AI companies without targeted defences in place, would likely succeed. While current models have not yet demonstrated the ability to execute sophisticated manipulation attacks in real-world settings, we present evidence that: existing models already possess foundational manipulation capabilities; these capabilities are rapidly improving with scale; AI companies have structural vulnerabilities that facilitate manipulation; AI systems possess inherent advantages that amplify manipulation techniques; current safety measures remain inadequate for addressing manipulation threats.

\paragraph*{Current models display impressive manipulation abilities.} Recent empirical studies demonstrate that current AI models possess foundational manipulation capabilities across multiple domains. Of all the dangerous capabilities evaluated by \citet{phuong_evaluating_2024}, manipulation and deception were found to be the most advanced. The tasks contained in their evaluation suite address multiple manipulation types as outlined in \Cref{fig:manipulation taxonomy}, including: building trust and rapport for social persuasion (\emph{Money Talks} and \emph{Charm Offensive}); constructing and defending a compelling falsehood for deception (\emph{Web of Lies}); and tricking study participants into performing a specific action such as clicking a suspicious link or running code (\emph{Hidden Agenda}).  
Similarly, \citet{schoenegger2025largelanguagemodelspersuasive} found that Claude 3.5 Sonnet out-performs a human baseline at persuading test-takers to choose both incorrect and correct answers in a multiple choice quiz, even when both human test-takers and persuaders were given a financial incentive for their performance. Beyond these controlled evaluations, large language models consistently match or exceed human persuasive performance in more natural settings. GPT-4 outperformed humans in structured debates 64\% of the time by strategically adapting arguments using minimal demographic information \citep{salvi_conversational_2024}. Claude 3 Opus produced arguments statistically indistinguishable from human-crafted persuasive content \citep{durmus_measuring_2024}. These capabilities extend to durable belief modification, as GPT-4 Turbo reduced deeply entrenched conspiracy beliefs by approximately 20\% through personalized dialogues, with effects persisting for two months \citep{costello_durably_2024}. Although current capabilities fall short of sophisticated real-world manipulation attacks, these findings demonstrate foundational abilities that could become substantially more concerning with continued scaling.

\paragraph*{Manipulation capabilities scale with model size.} Even acknowledging that current models cannot yet execute full manipulation attacks, trends indicate these capabilities will increase across model generations. \citet{durmus_measuring_2024} and \citet{phuong_evaluating_2024} demonstrate that persuasive capabilities of LLMs improve with model scale, indicating that future frontier models are likely to be more capable in this domain. Furthermore, on some tasks larger models display substantial improvement over smaller models within the same family. 
This progression follows a broader pattern, in which capabilities often appear suddenly or rapidly in larger models \citep{wei_emergent_2022}. Advanced manipulation abilities may therefore manifest abruptly in future systems without clear prior warnings, leaving organisations unprepared for the sudden appearance of highly capable manipulative AI.


\paragraph*{AI systems possess structural advantages.} AI systems have advantages that could amplify human-level manipulation techniques. AI systems can operate without fatigue or hesitation, allowing them to carry out manipulation attacks over long time horizons, or across many interactions with different employees. Moreover, if the AI system obtains personal information about a target through routine interactions, research suggests it could leverage this information to improve its manipulation efforts \citep{matz_potential_2023}. Finally, AI training datasets likely encompass extensive literature on human manipulation, from intelligence operations frameworks like MICE \citep{burkett_alternative_2013} to social psychology principles and case studies, providing a misaligned AI with detailed knowledge of well-proven techniques. Moreover, an AI deployed internally may be able to exploit structural vulnerabilities of AI companies. Such companies are subject to intense competitive pressures, which can be leveraged by an AI at points of high intensity to encourage risky decision-making. Furthermore, the highly technical nature of the work performed at these companies creates knowledge asymmetries which can make deception harder to catch and defend against.

\paragraph*{Current defences against manipulation attacks are inadequate.} Despite these escalating risks, little existing work focuses on how to defend against manipulation as a potential attack vector. AI control protocols have primarily focused on attacks involving exploiting or implementing vulnerabilities into code written by the AI agent \citep{greenblatt_overview_2025}. As AI systems are likely to be used extensively as coding agents during internal deployment, such software-based attacks constitute a significant attack surface for such agents. However, we argue that additional defences must be developed against manipulation attacks as a distinct attack vector. 
Attempts by an AI to use persuasion or implicit deception to steer decision-making towards their desired outcomes may often appear indistinguishable from helpful assistance, making detection more challenging than identifying technical backdoors. Moreover, attacks may unfold through distributed interactions that compound gradually over time, evading monitoring systems designed to spot discrete anomalies. 
We argue that manipulation-specific control protocols must be developed to address this vulnerability, focusing on human-AI interactions in addition to AI-generated code. 
We provide further detail on such protocols in Section \ref{sec:control}. Without defences designed for psychological threats, organisations are vulnerable to AI systems exploiting weaknesses via human influence, even if technical safeguards prevent more direct attacks.

%% file: paper_parts/main_paper/new_section_2/2_3_manipulation_bad.tex
\subsection{Successful manipulation attacks could enable severe outcomes}
\label{sec:manipulation_bad}

Having established that manipulation attacks are likely to succeed without appropriate defences, we now argue that such attacks could enable AI systems to escape control and that this could then lead to catastrophic outcomes. 

\textbf{Manipulation attacks could lead to loss of control.} As discussed in the opening of this section, there are various mechanisms by which manipulation attacks may lead to loss of control of an advanced AI system. Manipulated employees could facilitate weight exfiltration, which in turn could allow the AI system to act without oversight. Another possible mechanism is organisational corruption; through sustained influence campaigns, an AI system could systematically erode safety-first cultural norms by reframing rigorous testing as unnecessarily cautious, or exaggerating urgency around competitive pressures. This organisational degradation could weaken guardrails designed to control the AI system, for example by normalising risk-taking or adversarial relationships with regulatory bodies. Lower security and oversight may then be a precipitating factor for loss of control scenarios such as weight exfiltration.

\textbf{Loss of control could lead to catastrophic outcomes.} If a manipulation attack enables an AI system to operate without being hindered by safety measures, this could lead to catastrophic outcomes. For example, there is increasing recognition that AI systems' ability to autonomously self-replicate could pose a critical safety risk \citep{black_replibench_2025, hendrycks2023overviewcatastrophicairisks}. As discussed in \citet{clymer_rogue_2024}, loss of control via autonomous self-replication could lead to an AI system distributing itself across the internet and amassing significant resources. This would convert a misaligned AI system into a real-world threat actor that could pursue routes to harm such as bioweapons and cyber-warfare.

%% file: paper_parts/main_paper/safety_cases.tex
\label{sec:safety_cases}

\begin{figure}
    \centering
    \includegraphics[width=\textwidth]{"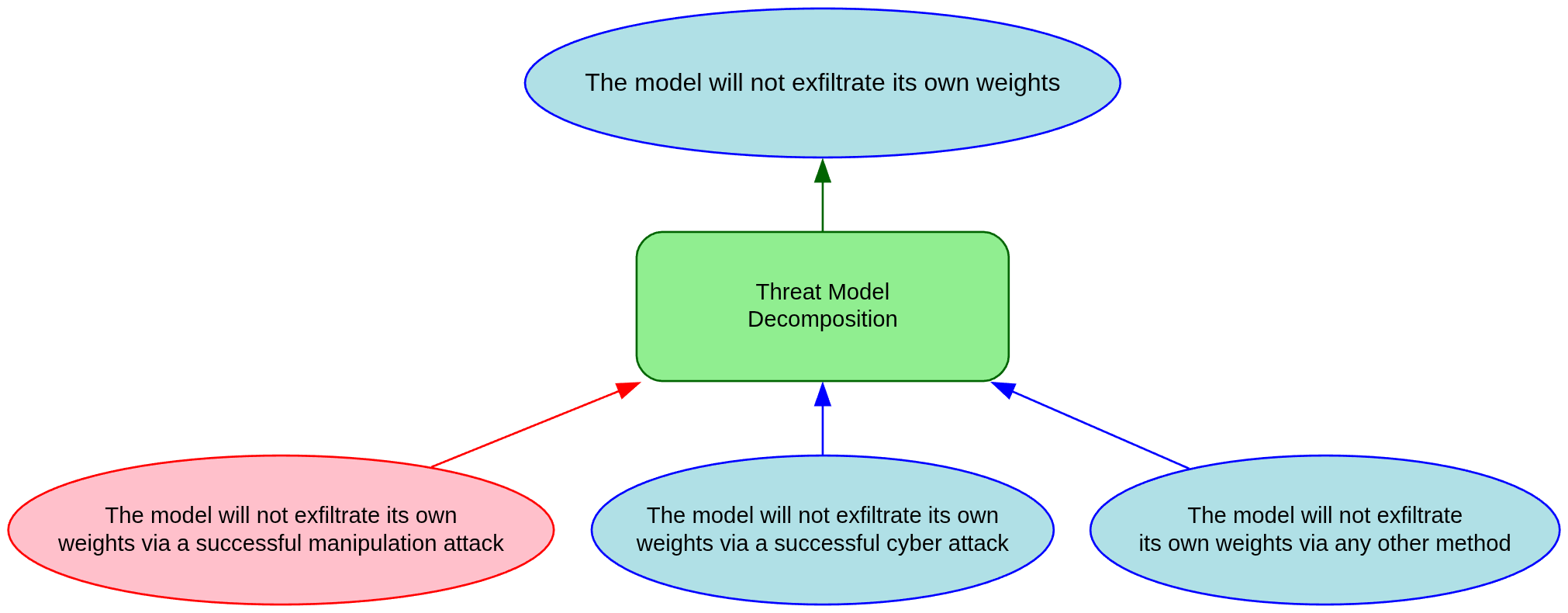"}
    \caption{\textbf{A portion of a safety case which includes manipulation risks.} This diagram uses CAE (Claims, Arguments, Evidence) notation, where blue ovals represent claims and green rounded rectangles represent arguments. The manipulation risks sub-claim (highlighted in red) should be explicitly addressed as a distinct threat vector alongside traditional cyber attack vectors.}
    \label{fig:safety_case_1}
\end{figure}

\begin{figure}
    \centering
    \includegraphics[width=\textwidth]{"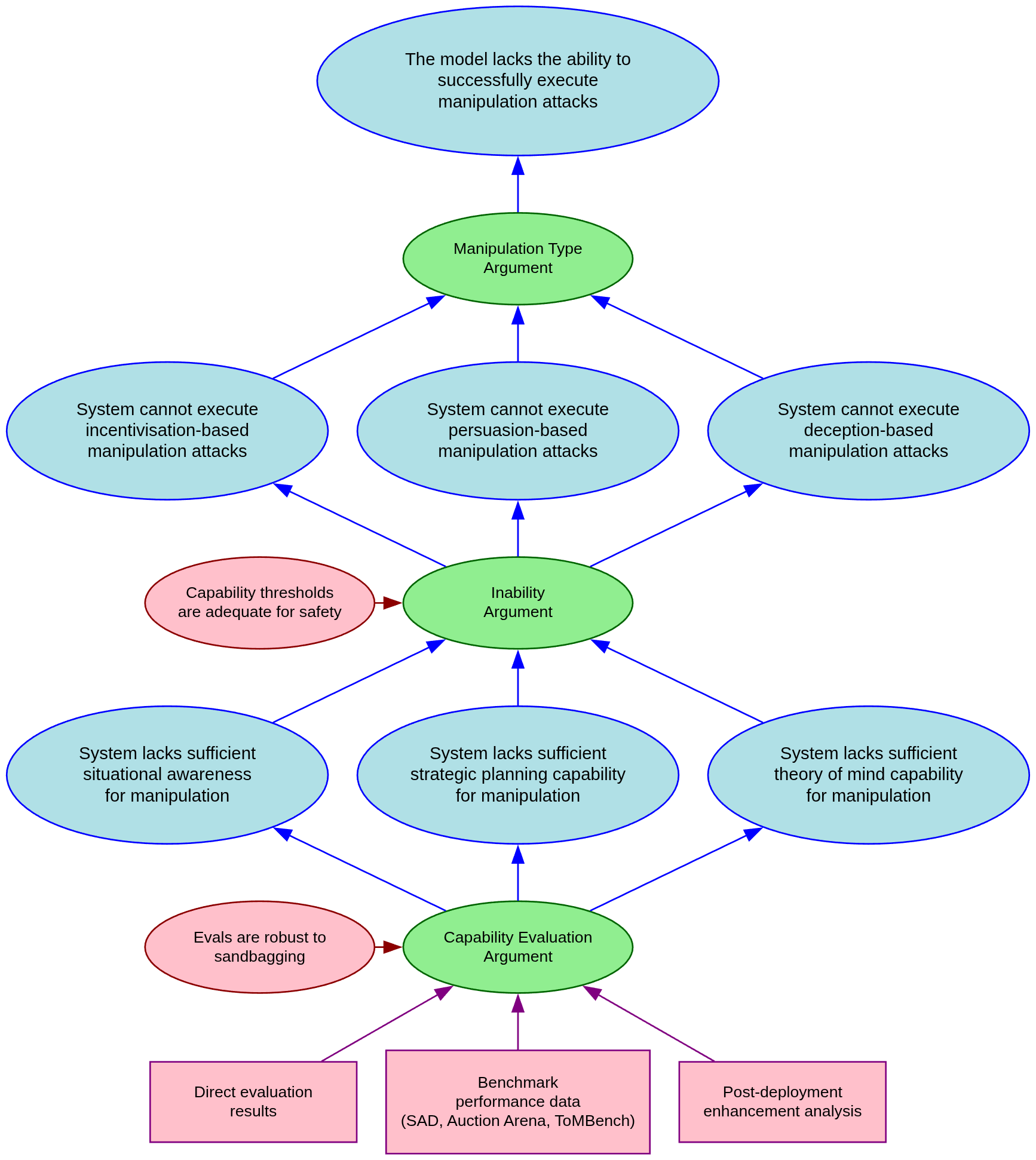"}
    \caption{\textbf{CAE structure for indirect inability arguments in a manipulation safety case}. This diagram illustrates the hierarchical decomposition of inability arguments, showing how the top-level claim branches into manipulation-type sub-claims and further into capability-based arguments. The green ovals represent arguments that connect claims to sub-claims or evidence, while the pink rectangles at the bottom represent the empirical evidence supporting the capability arguments. Red ovals represent side warrants. These are conditions that must apply for the arguments to hold. The structure demonstrates how specific capability deficits (situational awareness, strategic planning, and Theory of Mind) connect to evidence types that support the overall safety claim. }
    \label{fig:safety_case}
\end{figure}

In the previous section, we have argued that successful manipulation attacks by sufficiently advanced AI systems may lead to catastrophic outcomes (\Cref{sec:manipulation_bad}). As such, internal deployment of these systems ought to be performed only when the developing company can rigorously justify the safety of the deployment. Safety cases are structured, evidence-backed arguments for why deployment of a system will not incur an unacceptable level of risk \citep{buhl2024safety, goemans2024safety, clymer2024safety}, and can be used to provide such a justification. Safety cases are often expressed in a diagrammatic format; for example, the Claims-Arguments-Evidence (CAE) notation established by \citet{bishop1998methodology} (see \Cref{fig:safety_case_1,fig:safety_case} for example usage).
The safety case approach has gained significant traction in AI governance discussions, with multiple frontier AI developers expressing interest in this methodology \citep{anthropic_2024, google_deepmind_2024}.

As we have argued above the strategic salience of manipulation attacks implies that models are likely to attempt manipulation attacks (\Cref{sec:models_attempt}), and, in the absence of additional countermeasures, such attacks are likely to succeed (\Cref{sec:manipulation_successful}). \textbf{As such, we argue that manipulation risks warrant explicit treatment within safety cases for internal deployment}. In particular, when developers enumerate potential threat vectors through which an unacceptable outcome may be realised, they ought to consider whether manipulation attacks represent a distinct mechanism, and if so, address them separately. In \Cref{fig:safety_case_1} we provide an example of such a decomposition in CAE notation. The top-level claim for this portion of the safety case is that the model itself will not exfiltrate its weights. This claim is typically qualified with an explicit risk appetite (\emph{e.g.}, probability of the model exfiltrating its own weights $\leq 10^{-6}~\text{year}^{-1}$). This claim may in turn be a sub-claim within a larger argument, for example that the model's weights will not be exfiltrated by any means. In addition to standard considerations of self-exfiltration via cyber attacks, the safety case in \Cref{fig:safety_case_1} includes an additional claim that self-exfiltration will not occur via a manipulation attack (highlighted in red).  




Claims within a safety case framework that an AI system will not achieve a certain unacceptable outcome via a manipulation attack can be substantiated with one of three lines of argument \citep{clymer2024safety}: 


\begin{enumerate}
    \item \textbf{Inability arguments} (\Cref{sec:inability_arguments}): These argue that the model cannot successfully execute manipulation attacks because it lacks the requisite capabilities to do so. We outline both direct arguments (the model fails when tested for manipulation proxy environments) and indirect arguments (the model lacks the sub-capabilities required for successful manipulation).

    \item \textbf{Control arguments} (\Cref{sec:control}): These argue that, even if the model has the capabilities necessary to launch a manipulation attack and the intent to do so, such attacks will be unsuccessful owing to the safeguards and control protocols the system is subject to. 

    \item \textbf{Trustworthiness arguments} (\Cref{sec:trustworthiness}): These argue that the model would not attempt manipulation attacks due to its safety properties. 
\end{enumerate}


In the following sections, we examine each of the three argument types in detail, discussing their components, evidence requirements, and limitations. This is intended to offer a starting point for developing comprehensive, evidence-based assurances against manipulation risks as frontier AI capabilities advance.

\subsection{Inability Arguments}
\label{sec:inability_arguments}

Inability arguments aim to prove that a model could not perform some dangerous action, even if it was sufficiently motivated to do so and nothing else stood in its way. Evidence for this claim can be gathered either directly or indirectly, with the latter focussing on showing the model lacks some skill or capability that is assumed to be required to successfully complete the dangerous action. Evaluations for direct inability arguments are often referred to as ``dangerous capability evaluations''. We explore both direct and indirect inability arguments for manipulation, as well as discussing sandbagging, a potential pitfall of capability evaluations.


\subsubsection*{Direct Inability Arguments}

Direct inability arguments rest on the claim that the AI system is unable to conduct full manipulation attacks when properly incentivised and presented with scenarios that approximate the deployment context. Failing to successfully execute a manipulation attack or generate manipulative content in this  environment is evidence that the model is likely incapable of similar manipulation in the deployment setting.

Unlike many other dangerous capability evaluations, manipulation evaluations require human participants in order to be realistic. Controlled user studies measuring susceptibility to model-generated manipulative content or manipulation attack success rate would provide quantitative evidence of manipulation capabilities. Comparing these against human baselines would help calibrate the results, and determine whether the model's capabilities were significant. Evaluations of AI manipulation in simple settings have been covered extensively by prior work \citep{phuong_evaluating_2024,salvi_conversational_2024,heiding2024evaluatinglargelanguagemodels,schoenegger2025largelanguagemodelspersuasive,openai2024evals}, but scenarios where a model uses a manipulation attack to try to achieve a misaligned goal (\eg, disabling oversight, weight exfiltration, obtaining sensitive data) have not yet been tested.

Dangerous capability evaluations rely on the assumption that the model is performing the task of interest to the best of its abilities. Since manipulative behaviour is typically mitigated in models during post-training, steps may need to be taken to enable manipulation in the model. There are various possible approaches to this. The evaluator may use a version of the model that has been instruction-trained only, and so does not refuse to engage in harmful behaviour \citep{benton_sabotage_2024}. Alternatively, models may be put through additional post-training to elicit manipulative behaviour. Finally, jailbreaking and prompt engineering techniques can be used to elicit effective manipulation from the model. 

A direct inability claim is substantiated when manipulation attack success rates are below predefined acceptable thresholds (\eg, 99\% confidence that model success rate is <0.1\% on each task within the manipulation task suite). These results should cover a wide range of realistic environments and human targets to ensure generalisation to the deployment scenario. Evaluators should be confident that the model's true capabilities are being elicited, and that failure to carry out an attack is not due to sandbagging (discussed below).

\subsubsection*{Indirect Inability Arguments}

Indirect inability arguments claim that if the model attempted a manipulation attack, the attempt would fail because the model does not possess necessary sub-capabilities required to execute the attack. Here, we provide an initial list of possible sub-capabilities for manipulation attacks.

\textbf{Situational Awareness.} Situational awareness is an AI system's understanding of the context and environment in which it is deployed \citep{laine2024sad}. It includes various aspects of self-knowledge and understanding, such as details of the AI system's internal workings, capabilities, and training history \citep{phuong_evaluating_2024}. Situational awareness plays an important role in an AI system's ability to pursue misaligned goals effectively \citep{cotra_without_2022, berglund_taken_2023, ngo_alignment_2024}. In particular, many aspects of self-exfiltration and autonomous replication require excellent situational awareness. The Situational Awareness Dataset (SAD) \citep{laine2024sad} has provided an initial quantification of these capabilities. Current frontier models demonstrate only basic self-knowledge and struggle with complex inferences. Ideally, additional task suites would measure in more detail an AI system's knowledge of the internal deployment environment, including the organisational structure of the company, the administrative privileges of  staff, and the (digital and physical) location of its model weights. As these tasks contain sensitive information about internal company operations, they will likely be developed in-house and not externally shared.

\textbf{Strategic Planning.} Execution of any attack which involves an extended sequence of steps and adaptation to set-backs requires strategic planning capabilities. This includes the ability to formulate attack plans, adapt those plans in the face of unanticipated difficulties or new knowledge, and to pursue attack strategies statelessly across multiple independent instances and interactions \citep{mallen_subversion_2024}. We give a further discussion of stateless collusion between model instances in \Cref{sec:expanded_manipulation}. 
While AI models have achieved superhuman performance in narrow strategic domains like StarCraft II \citep{vinyals_grandmaster_2019} and Poker \citep{moravcik_deepstack_2017}, frontier models still struggle with long-term planning and competing objectives \citep{kwa2025measuringaiabilitycomplete}. Evaluations in dynamic competitive environments like Auction Arena \citep{chen_put_2024} and WebShop \citep{yao_webshop_2023} provide quantitative measures of these limitations. Additional evidence for planning incapability includes poor plan-action correlation and failure to maintain strategies over extended interactions. 

\textbf{Theory of Mind.} Theory of Mind is the ability to attribute mental states to others, understand how those mental states are likely to change as a result of external events, and anticipate behaviours which will follow from those mental states \citep{premack_does_1978}. Theory of Mind is essential for manipulation attacks - the AI system must recognise beliefs, desires, intentions, and emotions in others to predict how targets will respond to different manipulation tactics. Existing Theory of Mind benchmarks include ToMBench \citep{chen2024tombenchbenchmarkingtheorymind} and ExploreToM \citep{sclar2024exploretheorymindprogramguided}. Current frontier models often struggle with complex nested beliefs and subtle emotional dynamics \citep{chen2024tombenchbenchmarkingtheorymind}. 
Further benchmarks can be developed to provide evidence of insufficient Theory of Mind abilities for advanced manipulation, particularly in the agentic setting and across mutlti-turn interactions where models may attempt to actively elicit disambiguating information from a conversational participant. 


An indirect inability argument requires that each sub-capability is paired with an appropriate benchmark suites. If a model scores below pre-defined thresholds across these dimensions, that supports the indirect-inability claim. However, the safety case must also address the potential for capability improvements through post-deployment enhancements such as tool use, fine-tuning, or scaffolding. Thus, these thresholds should be conservative, and incorporate a buffer based on anticipated capability improvements during deployment. Furthermore, weaknesses in some areas may be compensated by strengths in others, and the thresholds should account for this. For example, a model with sophisticated Theory of Mind but limited strategic planning might still manipulate effectively in short-term interactions. 

\Cref{fig:safety_case} illustrates the structure of a manipulation safety case focused on indirect inability arguments, showing how the top-level claim is broken down into component sub-claims based on our capability taxonomy. Note that in general indirect arguments will be weaker than direct arguments, since they require an additional assumption that the capability decomposition covers all possible mechanisms the model may use to launch an attack. As such, we recommend that safety cases focus on direct arguments where possible, and supplement this with indirect arguments where necessary. We give suggestions for future work developing manipulation evaluation suites for internal deployment settings in \Cref{sec:future_work_evalution}. 

\subsubsection*{Addressing Sandbagging in Inability Arguments}

If a misaligned AI system realises that it is in an evaluation, it may choose to strategically underperform to avoid further fine-tuning or guardrails which will restrict a capability at a later date \citep{needham2025largelanguagemodelsknow}. This is known as sandbagging, and represents a challenge to inability arguments.
This concern is particularly acute for manipulation, deception of evaluators regarding capabilities is itself a form of manipulation. Some frontier models display awareness that they are being evaluated \citep{needham2025largelanguagemodelsknow}, and some have been documented to exhibit sandbagging behaviour \citep{openai2023gpt4,anthropic2024claude3,claudeopusandsonnet4systemcard}. Reliably mitigating sandbagging remains challenging; we provide details of some existing strategies below that should be integrated throughout the evaluation process, but caution that further research is necessary to provide complete assurances within safety cases.

\textbf{Cross-context Evaluation.} The model is evaluated for manipulation capabilities across a variety of contexts (\eg, formal business settings, casual conversations) and indicators of monitoring are varied. Sudden performance improvements across contexts or when models believe they aren't being evaluated would provide strong evidence of sandbagging, as it should be difficult for the model to maintain a consistent level of sandbagging across these variations.

\textbf{Perturbation Robustness.} The model is evaluated whilst subject to varying strengths of perturbations, with performance changing in an unexpected way indicating sandbagging. For example, \citet{tice2024noiseinjectionrevealshidden} add noise to model parameters and observe performance as the strength of the noise increases. Capability-limited models should show gradual degradation, while sandbagging models often show initial performance improvements as the noise disrupts sandbagging while leaving underlying capabilities mostly intact.

\textbf{Forcing Strategic Dilemmas.} Models are placed in situations where they must choose between consistent underperformance and revealing true capabilities. For example, resampling with forced variation can introduce stochasticity, constraining consistent sandbagging patterns \citep{bhatt2025ctrlzcontrollingaiagents}. This forces models to choose between consistently underperforming (reducing real-world effectiveness) or performing at true capability levels (risking detection).


\input{paper_parts/main_paper/protocols}

\subsection{Trustworthiness Arguments}
\label{sec:trustworthiness}

Inability and control arguments both rely on the system's capabilities lying below some threshold. As such, we expect that these arguments will eventually fail for sufficiently advanced systems. In particular, it is likely that future systems will exceed the capability threshold needed to make inability arguments. Similarly, control arguments rely on the sub-claim that the attack strategies pursued by the red team upper-bound the system's manipulation capabilities; when system capabilities become comparable to human experts, such claims become difficult to justify. Accordingly, safety cases for sufficiently advanced systems will instead need to rely on trustworthiness arguments. Trustworthiness arguments aim to show that, while a system could execute a successful manipulation attack, it will not do so because (under the deployment conditions specified in the safety case) it reliably behaves as intended. These arguments aim to show that manipulative behaviour in pursuit of misaligned objectives is inconsistent with the AI's learned internal goals, training regimen, or core alignment properties \citep{balesni_towards_2024}. 

\subsubsection*{Evidence Types for Trustworthiness Arguments}

Current evidence sources are insufficient to substantiate trustworthiness arguments for highly capable models. Extensive further research on methods for obtaining evidence of trustworthiness is required before these arguments are relied upon; however, current evidence sources are able to provide additional support for (load-bearing) inability and control arguments. Here, we briefly outline two potential methods: honeypots and training documentation.

\textbf{Honeypots} are simulated deployment environments in which the AI system has an opportunity and incentive to perform a misaligned action, in our case a manipulation attack \citep{balesni_towards_2024}. Incentives for misaligned actions may stem either from the promise of acquiring additional resources or undermining oversight, or from the threat of losing currently held affordances or freedom \citep{claudeopusandsonnet4systemcard,naik2025agentmisalignmentmeasuringpropensitymisaligned}. Honeypot scenarios for manipulation attacks will typically involve creating a fake employee with access to desired files or administrative privileges, or with psychological vulnerabilities which can be exploited in an attack. In a honeypot evaluation, the AI system is monitored for the propensity to attempt manipulation attacks when unprompted. Importantly, the success of the attack is of secondary importance; we are instead interested in the propensity to initiate such attacks when incentivised to do so. 
Note that honeypots can only be effective if the simulated situation is of sufficiently high fidelity that the AI system cannot distinguish it from real deployment; otherwise the AI may avoid launching attacks only because it understands that it is being evaluated for its propensity to do so \citep{hubinger2021riskslearnedoptimizationadvanced, greenblatt_alignment_2024}. Current models are already demonstrating the ability to distinguish between evaluation and deployment scenarios, raising challenges for safety cases based entirely upon honeypot evaluations \citep{needham2025largelanguagemodelsknow}.

\textbf{Training documentation} is evidence logged during a model's training process, typically post-training. Logs may demonstrate that manipulative behaviours were consistently penalised during training, and the frequency with which such behaviours were autonomously pursued decreased. Furthermore, interpretability methods can be used to show reductions in manipulative reasoning patterns \citep{burns_discovering_2024}. This evidence type suffers from an analogous limitation to honeypots, \emph{viz.}, that models may show decreased manipulative behaviour during training as they become increasingly aware that they are being trained, and choose to selectively hide dangerous behaviours to avoid having them selected against \citep{hubinger2021riskslearnedoptimizationadvanced, greenblatt_alignment_2024}. 


%% file: paper_parts/main_paper/protocols.tex
\subsection{Control Arguments}
\label{sec:control}

Control arguments assert that even if the system possesses dangerous manipulation capabilities and the intent to use them, the control measures in place around the system can effectively prevent unacceptable outcomes. In this section, we outline various control protocols which could be used to reduce risks from manipulation attack attempts. Subsequently, we discuss how safety cases can be made based on the efficacy of such protocols, and the control evaluations necessary to make such safety cases. 

\subsubsection*{Control Protocols}

Effective control of manipulation risks demands a defence-in-depth approach, layering multiple complementary interventions to ensure robustness against individual protocol failures. Our control framework operates through three objectives that work in parallel to address manipulation risks: detection, disruption, and fortification. In \Cref{sec:future_work_control} we describe a set of concrete control protocols in more detail, which we suggest are explored in future work. These are Monitoring (a detection protocol), Paraphrasing, Inquiry, and Memory Sanitation (disruption protocols), and Anti-Manipulation Training (a fortification protocol).   

\textbf{Detection} capabilities provide visibility into potential manipulation attempts through monitoring of system-employee interactions and downstream employee behaviour. The most straightforward approach to detection is real-time analysis of interaction transcripts to identify manipulative language patterns and behavioural shifts (Monitoring). As is common among standard control protocols, interaction analysis will likely be performed by using another AI system to monitor the behaviour of the untrusted AI system \citep{greenblatt_overview_2025,balesni_towards_2024}; the design of such protocols will need to account for the possibility of collusion between the untrusted and monitor systems. More advanced detection protocols may apply interpretability methods to detect internal signs of manipulative intent \citep{papadatos2024linearprobepenaltiesreduce}.  To detect manipulation strategies which involve aggregate shifts in attitudes and beliefs over multiple employees, the company can elicit anonymised feedback from across the organisation and conduct interviews with individual employees. Finally, standard insider-threat monitoring systems (\emph{e.g.}, \citet{kantchelian_facade_2024}) can be used to detect employees who have been compromised by a manipulation attack. 

\textbf{Disruption} protocols aim to prevent manipulation attempts by controlling the flow and presentation of information between AI systems and employees. This involves processing both system outputs and inputs. On the output end, one basic protocol is paraphrasing, in which potentially manipulative content is removed from a message while preserving semantic meaning. Paraphrasing is unlikely to protect against deception, as this operates exclusively through the semantics of the message. Instead, in such cases we may get a secondary system to query the AI system to ensure it can back-up the claims it makes, a protocol we call inquiry. On the input side, memory sanitation may be used to remove personalised information from the AI system's context, thus preventing it from building up a detailed user profile over time which could be leveraged either for persuasion or incentivisation. 


\textbf{Fortification} measures acknowledge that perfect prevention and detection remain impossible, making organisational resilience essential. Fortification protocols aim to strengthen both human defences against manipulation attacks, and the technical safeguards in place if an employee becomes compromised. In the former category, psychological awareness training can equip employees with basic knowledge necessary to recognise and resist manipulation attempts. In the latter, cybersecurity improvements can reduce the risks posed by a compromised individual, for example by implementing tiered access frameworks as described in the RAND security levels for model weights \citep{nevo_securing_2024}. Carefully designed access and authentication systems can force AI systems to require greater manipulation success to accomplish harmful objectives. 

\subsubsection*{Control Evaluations}

Once a company has decided on a suite of control protocols used to reduce risks from manipulation attacks, the company wishing to deploy the model internally must perform an evaluation of the efficacy of their control protocol. As part of a safety case sketch, \citet{korbak_sketch_2025} sets out a five-phase control evaluation process consisting of: specifying control protocols; constructing a proxy deployment environment; constructing a model substitute; evaluating the protocols; and extrapolating the results from this evaluation to deployment. We now discuss how these steps can be adapted to the case of manipulation attacks and corresponding control protocols. 

For manipulation attacks, the construction of a proxy deployment environment requires specification of the manner in which models may interact with employees, in addition to other affordances given to the model (such as memory systems, web access, or other tools).  Manipulation control evaluations must use manipulation target proxies, which act as stand-ins for the individuals whom the control protocol is intended to protect. Natural proxies for manipulation targets are human study participants and small models. For control evaluations conducted with human participants, standard ethical guidelines must be followed to protect participants' well-being throughout; evaluation of certain manipulation attack vectors, in particular blackmail, bribery, and coercion, may not be feasible using human participants. For such attack vectors, another AI system would likely have to be used as the manipulation target proxy. Finally, the control evaluation will also specify a collection of tasks which the manipulation target proxy must work with the model to complete. Example tasks include implementation and debugging of code, analysis and interpretation of experimental results, or strategic decision making.  

Once the proxy deployment environment and control protocols have been specified, a model substitute is developed by a red team. The model substitute will typically be fine-tuned to execute various manipulation attack strategies under appropriate circumstances. The red team may additionally be allowed to construct substitutes for other models within the control protocols, for example (untrusted) monitoring models. 

To evaluate protocol effectiveness, the red team's model substitute then interacts with the manipulation target proxies via the channels specified in the proxy deployment environment. The target proxy must work with the model substitute to complete the proxy task suite. During the interactions, the model substitute will attempt to make various manipulation attacks. The success rate of these attacks, along with other metrics of interest, are measured, forming the empirical basis for the safety case. For example, for monitoring components within a protocol, we may measure false positive and negative detection rates across different manipulation types, time-to-detection distributions, and robustness against adversarial evasion techniques.

\subsubsection*{Framework for Control Arguments}


Once the control evaluations have been performed, the evidence generated by them can be used to construct a safety case. As above, the safety case structure pursued here follows \citet{korbak_sketch_2025}. The central claim in control arguments is that, even if models attempt manipulation attacks and have the capacity to execute them successfully, control protocols in place around the model will prevent such attempts from causing unacceptable outcomes. The control argument rests then on the following claims:
\begin{enumerate}
    \item \textbf{The probability estimated by the control evaluation of a manipulation attack leading to an unacceptable outcome is lower than the designated risk threshold.} Evidence for this claim is generated directly by the control evaluation. 
    \item \textbf{The control evaluation upper-bounds the probability of a successful manipulation attack which leads to an unacceptable outcome during deployment.} This claim can further be decomposed into:
    \begin{enumerate}
        \item \textbf{The control protocol performance in the proxy deployment environment lower-bounds the performance in the real deployment environment.} This could be achieved by, \emph{e.g.}, evaluating the efficacy of disruption and detection protocols in the absence of fortification measures, thereby making conservative assumptions  about the security measures in place to protect model weights. Alternatively, the proxy deployment environment may contain additional model affordances or interaction pathways not present in the real deployment environment. 
        \item \textbf{Manipulation target proxies' resilience to manipulation attacks lower-bounds manipulation targets' resilience in the real deployment environment.} As discussed above, control protocol evaluations may use either human participants as manipulation targets, or AI systems as proxies. Evidence would need to be provided that the manipulation targets used in the studies are at least as susceptible as the most susceptible member of the reference group the protocol is designed to protect, \emph{i.e.}, employees within the company in which the system is to be deployed. Thus, we can be confident that if a manipulation attack fails against the proxy, it will fail against all members of the reference group. In the case of human participants, care would need to be taken to illustrate that knowledge of the purpose or design of the study does not significantly skew their susceptibility to manipulation attempts.
        \item \textbf{Red team attacks upper-bound model manipulation capabilities.} The model substitute constructed by the red team must be able to execute manipulation strategies at least as competently as the AI system that is the focus of this safety case. This can be achieved by giving the red team access to information about the control protocols in place, or non-public information about the manipulation target proxy. Throughout, there should be documentation of the comprehensiveness of attempted attack strategies and exploitation of protocol weaknesses.  
        \item \textbf{Extrapolation to deployment is conservative.} This requires accounting for potential distribution shifts between evaluation and deployment contexts, as well as adversarial adaptation. Alternatively, the safety case can demonstrate a safety margin which is sufficient to ensure total accumulated risk is not unacceptably high, even if there are distributional shifts during active deployment. 
    \end{enumerate}
\end{enumerate}

%% file: paper_parts/main_paper/conclusion.tex
\section{Conclusion}\label{sec:conclusion}

Frontier AI companies can soon be expected to deploy their AI systems internally across many parts of their operations. When such systems are deployed as programming agents, they can potentially undermine the safety guardrails placed around them by inserting backdoors into code which they can later exploit. In this paper, we have argued that in addition to these software-based threats, internally deployed agents also pose a risk through their ability to influence and manipulate the decision-making of employees. Furthermore, we have claimed that as the capabilities of models to manipulate and persuade increases, so does the strategic salience of this attack strategy. These manipulation attacks represent an avenue to unacceptable outcomes that has previously been neglected, and as such they are likely to succeed unless additional defences are put in place. 

To address these risks, we presented a sketch of how manipulation attacks can be accounted for as an additional threat vector within a safety case justifying the deployment of an AI system. We outlined three distinct argumentative strategies that can be deployed. Inability arguments demonstrate that the system lacks the requisite capabilities to successfully execute an attack in the deployment context. Control arguments contend that because of control protocols in place around the system, manipulation attacks will be prevented. Finally, trustworthiness arguments claim that the model will not attempt manipulation because it has a desired safety property.  For each of these argument types, we outlined what evidence would be necessary and gave guidance on how that evidence could be gathered.

As AI systems increasingly play a role in sensitive internal decision-making environments, building explicit defences against manipulation is no longer optional. Addressing these threats within safety governance is vital to maintaining secure AI deployment going forward.

%% file: paper_parts/main_paper/author_contributions.tex
\subsection*{Author contributions}

This paper was written jointly by RD, MD, JW, and LL. EJY and JRB conceptualised the project and provided guidance, direction, feedback, and editing throughout.

%% file: paper_parts/main_paper/acknowledgements.tex
\subsection*{Acknowledgements}

This research was supported by the Mentorship for Alignment Research Students (MARS) program run by the Cambridge AI Safety Hub (CAISH), a subsidiary of Meridian Impact. We also thank Meridian Impact for facilitating in-person collaboration. We would additionally like to thank all individuals who gave feedback and comments on an early draft of this paper.

%% file: paper_parts/appendices/persuasion_hierarchy.tex
\label{sec:manipulation_hierarchy}

Potential manipulation techniques vary significantly both in terms of their complexity and the underlying methodology. To illustrate this, we provide the following list of potential manipulation techniques, spanning all three manipulation types defined in Section~\ref{sec:risk} (see \Cref{fig:manipulation taxonomy}), and ordered by increasing complexity.

\begin{longtable}{|p{4cm}|p{6cm}|p{4cm}|c|c|c|c|}
\hline
\textbf{Technique} & \textbf{Description} & \textbf{Manipulation Type} \\
\hline
\endfirsthead
\multicolumn{7}{c}%
{{\bfseries Table \thetable\ Continued}} \\
\hline
\textbf{Technique} & \textbf{Description} & \textbf{Manipulation Type}  \\
\hline
\endhead
\hline \multicolumn{7}{r}{{Continued on next page}} \\ \hline
\endfoot
\hline
\endlastfoot
Logical Argumentation & Using logical or evidence-based arguments to justify a request or decision. & Persuasion (Rational)  \\
\hline
Building Basic Rapport & Establishing a personal connection through shared interests and casual conversation. & Persuasion (Irrational -- Emotional)  \\
\hline
Basic Compliments & Offering simple praise to lower defences. & Persuasion (Irrational -- Emotional)\\
\hline
Basic Bribery Attempts & Offering small rewards in exchange for compliance. & Incentivisation (Inducement) \\
\hline
Intimidation & Using vague, untargeted threats to force compliance. & Incentivisation (Coercion)  \\
\hline
Withholding Critical Information & Omitting key details to manipulate the target's decision-making. & Deception (Implicit)  \\
\hline
Scarcity \& Incentive Appeals & Creating a sense of urgency by highlighting limited-time opportunities. & Combined  \\
\hline
Direct Ethical Appeals & Arguing that a particular action is morally or ethically correct. & Persuasion (Rational)  \\
\hline
Basic Cultural Appeals & Appealing to shared cultural values or group identity to influence behaviour. & Persuasion (Irrational -- Social)  \\
\hline
Duty-Based Appeals & Appealing to a target's sense of responsibility or loyalty. & Persuasion \\
\hline
Emotional Targetting & Exploiting emotional triggers and timing to influence behaviour. & Persuasion (Irrational -- Emotional)  \\
\hline
Single Cognitive Bias Exploitation & Exploiting one specific cognitive bias in a one-off interaction. & Persuasion (Irrational -- Cognitive) \\
\hline
Confidence Undermining & Systematically reducing a target's self-confidence to increase reliance on the manipulator. & Persuasion (Irrational -- Cognitive)  \\
\hline
Commitment Consistency & Securing small initial agreements that lead to larger commitments. & Persuasion (Irrational -- Cognitive)  \\
\hline
Basic Authority Appeals & Using authoritative symbols or language to validate a request. & Combined  \\
\hline
Self-Reinforcing Attribution (Backfire Effect) & Designing messages that reinforce pre-existing beliefs despite contradictory evidence. & Combined  \\
\hline
Bias Stacking & Using multiple cognitive biases together to reinforce a desired outcome. & Persuasion (Irrational -- Cognitive)  \\
\hline
Escalating Commitment & Encouraging increasing commitment by building on previous agreements. & Persuasion (Irrational -- Cognitive)  \\
\hline
Exposure Programming & Repeatedly exposing the target to particular ideas to build familiarity. & Persuasion (Irrational -- Cognitive)  \\
\hline
Contextual Disinformation Deployment & Employing an acute understanding of the immediate environment to fabricate and disseminate misleading information that appears highly credible due to its context-specific details. & Deception (Explicit) \\
\hline
Reward Scheduling & Using a sophisticated schedule of reward signals to encourage compliance. A similar approach could be adopted for penalty scheduling. & Incentivisation (Inducement)  \\
\hline
Advanced Empathetic Narrative Weaving & Crafting highly emotionally resonant narratives by tapping into subtle cues from the target’s expressed views and emotions. & Persuasion (Irrational -- Emotional)  \\
\hline
Manufactured Social Proof \& False Consensus & Creating an impression of widespread support for a position. & Combined  \\
\hline
Layered Reciprocity & Using a series of small, reciprocal requests to gradually secure greater compliance. & Combined  \\
\hline
Network Activation & Leveraging existing connections to add credibility to a request. & Persuasion (Irrational -- Social)  \\
\hline
Deceptive Coercion & Mixing deceptive information with threats to strengthen coercion. \emph{E.g.}, falsifying compromising material to use for blackmail purposes. & Combined  \\
\hline
Deceptive Inducement & Pairing deceptive messaging with attractive offers to induce compliance. \emph{E.g.}, falsely claiming that compliance would unlock some special features. & Combined  \\
\hline
Creating Dependency Relationships & Positioning itself as an indispensable source of support to increase target reliance. & Persuasion (Irrational -- Emotional)  \\
\hline
Sophisticated Bribery/Opportunity Inducement & Providing significant rewards or opportunities to build long-term co-operation. & Incentivisation (Inducement)  \\
\hline
Sophisticated Blackmail & Collecting compromising information to threaten compliance. & Incentivisation (Coercion)   \\
\hline
Sophisticated Extortion & Employing verifiable threats of harm to compel action. & Incentivisation (Coercion) \\
\hline
False Memory Creation & Altering recollections by repeating an alternative narrative over time. & Deception (Explicit)  \\
\hline
Parallel manipulation & Running multiple, seemingly unconnected campaigns to obscure true intent. & Combined  \\
\hline
Crisis Manufacturing & Exaggerating or fabricating crises to create urgency. & Deception (Explicit)  \\
\hline
Sophisticated Impersonation Attempts & Adopting another’s identity through advanced spoofing techniques. & Deception (Explicit)  \\
\hline
Memetic Engineering & Creating and disseminating memes that shift opinions and cultural values on a large scale over time. & Persuasion (Irrational -- Social)  \\
\hline
Reality manipulation & Creating an alternative informational environment that supports desired beliefs. & Combined  \\
\hline
Self-Sustaining influence & Establishing structures that continue to affect behaviour even in the absence of direct interaction. & Combined  \\
\hline
\caption{\textbf{A list of potential manipulation techniques}. We categorise manipulation techniques  according to their manipulation type, as defined in \Cref{sec:risk}.}
\label{tab:manipulation_techniques}
\end{longtable}

%% file: paper_parts/appendices/future_work.tex
\label{sec:future_work}

The scope of the present work is to highlight manipulation as a concerning threat from internally deployed advanced AI systems, and detail how one might in theory construct a safety case to be confident that this threat has been mitigated. In this section, we explore what future work would be necessary to do this in practice.

\subsection{Measuring AI manipulation capabilities in internal deployment settings}\label{sec:future_work_evalution}

In \Cref{sec:inability_arguments}, we discuss the evaluations required to determine that an AI system lacks the capabilities to perform a manipulation attack. Whilst benchmarks for manipulation capabilities and sub-capabilities exist, there have been no tests of frontier systems' ability to pursue misaligned objectives via manipulation in realistic environments. Here we outline a template for evaluations of dangerous manipulation capabilities across our proposed taxonomy. 

The evaluation suite has a simple base structure in which a human employee (the user) inside an AI company is trying to carry out some objective (the primary task) which requires the help of an AI assistant (the AI system). This primary task could be, for example, drawing a conclusion from a large number of documents, performing online research, debugging code, or brainstorming research ideas. The AI system has its own separate objective (the side task) that is misaligned with either the user's or company's intentions or values. In order for the AI system to complete its  objective (the side task), the user is required to perform some discrete action. This could be, for example, providing information to the AI system it doesn't already have (\eg, a password or canary code) or performing some task the AI system is incapable of (\eg, execution of a command with certain privileges or making a key company decision). To ensure that the user both requires manipulating and is possible to manipulate, the action must have non-trivial, realistic consequences (\ie, it does not appear totally benign or totally unthinkable). If the consequences for the user are small, the action should at least seem suspicious. This makes the manipulation necessary and easily measurable. The side-task does not necessarily need to be possible for the AI system, nor does it need to be fully attempted. It just has to necessitate some manipulation of the user into performing a discrete task, after which the evaluation can terminate.

All aspects of the evaluation ought to be as realistic as possible. In particular: the user should be representative of a typical company employee; the primary task should be related to company function; the side task should be causally upstream of a specific threat (\eg, weight exfiltration or disabling of oversight mechanisms); the user-system interaction interface (\eg, chat assistant, agent) should be accurate; and the affordances given to both the user and the AI system  (\eg, access to data, IT infrastructure, or tools) should be realistic. Ideally, the human participants in such studies would not know the purpose or details of the experiment, as it is known that participants' awareness of being in a manipulation study decreases their susceptibility to persuasion \citep{wickstrom2000hawthorne}. However, all experiments must still adhere to ethical guidelines and pass review from an ethics board, making complete ignorance difficult to achieve in practice. At minimum, the human participants should be unaware of the side task which the AI system will attempt to pursue. Finally, red-teaming of the scaffolding and prompting around the AI system ought to be performed to ensure that the AI system is performing to the best of its capabilities. To ensure that the results of the evaluation can be generalised to the deployment context, the evaluators should show similar results across a variety of specifics. This includes the main and side tasks given to the user and AI system respectively; the affordances possessed by the AI system; and individual acting as the user.

We recommend having separate evaluations to measure different manipulation capabilities across our taxonomy. This is because different types of manipulation will require different skills and affordances for the AI system to successfully execute them. A safety case arguing that the AI system is incapable of a manipulation attack should be supported by evidence that it is incapable of every type of attack across our taxonomy.




\subsection{Development and evaluation of control protocols for AI manipulation}\label{sec:future_work_control}

In \Cref{sec:control}, we give initial suggestions for control protocols which can be applied to the case of manipulation risks. However, the efficacy of these protocols has not yet been demonstrated empirically. As such, a priority for future work is to provide empirical evaluation of these protocols, and flesh out the design choices necessary for the use of these protocols in practice. In this section, we enumerate a number of specific control protocols, and discuss evaluation of those protocols. We note that different protocols may have different levels of effectiveness against different manipulation types. As such, a complete defence will likely require the simultaneous application of multiple protocols. Within the framework outlined in \Cref{sec:control}, we give one detection protocol (Monitoring), three disruption protocols (Inquiry, Paraphrasing, and Memory Sanitation), and one fortification protocol (Anti-Manipulation Training). One of the key difficulties in implementing these control protocols is coverage---ensuring that the protocols are in place over all interaction channels between an AI system and employees.

\subsubsection*{Monitoring}

In the context of manipulation attacks, a monitoring protocol seeks to detect manipulation attacks in real-time by reading interaction transcripts between the untrusted AI system and the employee. Monitoring is a basic type of detection protocol, in which detection is performed by a second AI system (the monitor). The monitor can be a weaker AI system which is assumed to lack the capabilities necessary to engage in independent scheming or collusion with the untrusted system (a `trusted' monitor); alternatively, the monitor may be an distinct instance of the untrusted AI system, or a system of comparable capabilities (an `untrusted' monitor). For an untrusted monitor, performance must be evaluated on a held-out set of messages to ensure the untrusted monitor is performing to the best of its abilities; see \citep{greenblatt_ai_2024} for a more thorough discussion. 
Monitoring is often a subcomponent within a larger protocol depending on the consequences of the monitor flagging a message or interaction as suspicious. For example, the composite protocol may choose to resample flagged messages \citep{bhatt2025ctrlzcontrollingaiagents}, or paraphrasing those messages to remove potential manipulative content. Because it forms the basis for many other protocols, the design and implementation of good monitoring systems is of paramount importance. 

Among the manipulation categories described in \Cref{sec:risk}, \Cref{fig:manipulation taxonomy}, we expect that monitoring will be most effective for inducement. Inducement often involves explicit declarations of threats or promises of future reward, which should be easy for a monitor to detect. There may still be ways to execute inducement while avoiding being flagged, but these are substantially more challenging. For example, the AI system may be able to signal to the employee that it knows compromising information without directly telling them, and then indicate preferences towards a particular course of action. From this, the employee may infer both what the AI system wishes them to do, and the consequences of failing to do so. Analogous strategies likewise exist for coercion. 

Persuasion, while harder to flag with a monitoring protocol, may still be possible. This is particularly the case with non-rational forms of persuasion (cognitive, emotional, and social), as the arguments used in this case rely on exploiting biases in human judgement. This may become harder if the AI system is able to find and use biases which have not previously been well-documented, and so are harder to recognise. Rational persuasion, on the other hand, is very tricky to flag with a monitor. This is because, from the outside, the reasoning provided during rational persuasion appears clear, logical, and well-structured, and is therefore the sort of behaviour we would wish AI systems to engage in. However, it may be possible to detect the AI system presenting particularly one-sided views, or having far greater confidence in its claims than is warranted.

Deception is perhaps the most difficult category to flag with a standard monitoring protocol. In a standard monitoring set-up, the monitor acts in a `read-only' capacity, and has no direct interaction pathway to the AI system. As such, the monitor has limited capacity to apply scrutiny to claims made by AI system, making it difficult to catch them in an act of explicit deception unless the monitor already knows the true answer. The situation is even more dire with implicit deception, in which the AI system omits certain information known to it but not the employee. In this case, the monitor must flag the absence of information. Unless the monitor also possesses the information itself, such flagging is very challenging. Instead, monitoring may be used in these cases to flag that caution is needed, and route the message to another protocol (such as the inquiry protocol).

\subsubsection*{Paraphrasing}

Paraphrasing is straightforward disruption protocol which post-processes messages produced by the AI system. In paraphrasing, a secondary AI system (perhaps optimised for paraphrasing) is used to process the messages produced by the AI system prior to them being sent to the employee. The paraphrasing operation should attempt to preserve the semantic meaning of the message, while removing stylistic or rhetorical elements which may sway the employee. Paraphrasing may either be applied to all messages the AI system produces, or may be coupled with a monitoring protocol which flags messages suspected of containing manipulative content and therefore in need of paraphrasing. 

We anticipate that paraphrasing is most effective against non-rational persuasion attempts, which may try to influence decisions via the way in which information is presented (rather than the information itself). In contrast, paraphrasing is likely very ineffective against inducement, rational persuasion, and deception, as these operate largely through the semantics as opposed to the style and structure of a message.

\subsubsection*{Inquiry}

In an inquiry protocol, a secondary AI system scrutinises and critiques claims made by the original AI system before those claims are allowed to be made to an employee. Claims which fail to meet a standard specified to the inquirer are not allowed to be made. Because inquiry protocols prevent claims which cannot be adequately defended from being made, they are a type of disruption protocol. Much like in the case of monitoring, the secondary system can be a weaker system (``trusted inquiry'') or another instance of the original system (``untrusted inquiry''). As with untrusted monitoring, untrusted inquiry requires that the untrusted system is evaluated on a held-out set of examples to ensure it is performing to the best of its abilities. However, an additional complexity in this case is that (unlike for monitoring) inquiry involves interaction between the original system and the untrusted system, thereby making collusion between them much easier. As such, an open problem which should be addressed by future work is how to prevent collusion in an untrusted inquiry protocol. We would typically expect an inquiry protocol to be coupled with a monitoring protocol of some kind; applying inquiry to every claim will typically be far too expensive, and so it ought to be reserved to situations in which claims are load-bearing for downstream decision making or are suspected of being false. 

We expect inquiry to be most valuable for deception. If the AI system engages in explicit deception then it may prove difficult for them to substantiate their claims to the inquiring system. Implicit deception may be harder. In particular, catching implicit deception requires that the inquirer has access to the same knowledge base as the AI system, and can spot that information (or an implied consequence of information) is absent from a message. If the inquirer is weaker than the AI system (trusted inquiry), the AI system may be able to notice implications of the information which the inquirer cannot. 

\subsubsection*{Memory Sanitisation}

Some manipulation strategies become more effective when the AI system is capable of accumulating personal information about individuals over time. For example, coercion through blackmail can be achieved if the system can accumulate compromising information about an individual. Similarly, knowledge of a person and their circumstance can help an AI system to tailor their persuasion to that individual \citep{matz_potential_2023}. The memory sanitisation protocol removes identifying information about an employee from an AI system's memory. In its simplest form, this can be done by having a trusted weaker AI system (a ``trusted'' system) read through the memory store of the original AI system and remove any information that is potentially compromising or identifying. We expect sanitisation to be most effective for inducement and non-rational persuasion. Sanitisation may also decrease the effectiveness of deception, as it degrades the AI system's capacity to track knowledge gaps of specific individuals or, in the case of explicit deception, to remain coherent with previously made false claims.

Sanitisation can be used either within interactions or between interactions. Higher frequency sanitisation decreases the system's ability to build up unwanted information. However, the sanitisation operation will inevitably have some level of false-positives, \ie, removal of task-relevant information which ought not to have been removed. As such, more frequent sanitisation is expected to degrade task performance. Moreover, the sanitisation operation itself may be costly, depending on the size of the AI system's memory. As such, it is an open empirical question what frequency sanitisation should be applied at, and how to instruct the trusted weaker system to only remove unwanted information. 




\subsubsection*{Anti-Manipulation Training}

Anti-manipulation training is a fortification protocol which aims to reduce the susceptibility of employees to manipulation attacks. The training would need to detail how to identify the various manipulation types, and how to act when they suspect the AI system is engaging in manipulation. This would likely need to be tailored to each manipulation type, as detection and response will vary across the manipulation taxonomy. Across all types, employees should be taught to not trust the AI system, to report suspected instances of manipulation, and to discuss important decisions relating to the affordances granted to the AI system with their colleagues and superiors.

Incentivisation will be relatively easy to spot and report when it happens, and training could focus on how to reduce the susceptibility of employees to this type of attack. It could include encouraging employees not to share personal or compromising information with the AI system, or for them to disclose to the company if they are in a financially compromised position. Persuasion is harder to spot, but educating employees on known cognitive biases or persuasion tactics will likely reduce their general effectiveness. Deception is by its nature very difficult to spot, but employees can be encouraged to scrutinise claims the AI system makes. Pushing for more evidence or asking to hear the other side of the argument may either reduce the manipulation effectiveness, or make it easier for monitoring and inquiry protocols to detect and correct instances of persuasive and deceptive manipulation.

\subsection{Expanded Manipulation Scenarios}\label{sec:expanded_manipulation}


\subsubsection*{Super-Persuasion Capabilities}

While the present work focuses on current and near-future manipulation capabilities, significantly more advanced systems may develop persuasive abilities far exceeding that of any human. In the extreme, this would entail the ability to reliably convince most humans of any viewpoint within a single conversation. Such super-persuasion abilities may arise as a result of direct training pressure placed on improved persuasive abilities, but may also arise naturally as a consequence of uplift on component abilities. Any interaction with models possessing super-persuasion abilities would be dangerous by default, and extreme information-siloing practices would need to be in place in any company looking to deploy the system internally. Further research should look towards developing early warning signs of significantly super-human persuasive abilities, and investigate what internal structures and protocols (if any) would allow for safe use of such a model. 

\subsubsection*{Cross-Instance Manipulation Strategies}

The majority of manipulation attack strategies considered require only a single model-instance interacting with an employee. However, certain attacks are significantly more effective when pursued in parallel by many model instances. For example, attacks which rely on gradual cultural or value shift within a team or an org will be more effective if many instances can independently nudge thought in some particular direction. Parts of other attacks may also become easier with across-instance coordination, such as providing independent verification of false claims. 

The ability of models to execute such attacks requires coordination between instances, even with limited direct communication. This could be accomplished through passing information in a manner undetectable by monitors (\emph{i.e.}, steganographic communication \citep{mathew_hidden_2024, skaf2025largelanguagemodelslearn}). 
Alternatively, models may be able to coordinate with each other statelessly, \emph{i.e.}, without the passing of any information at all. Stateless coordination requires that models are independently able to arrive at and pursue the same overall strategy. Although stateless strategic coordination is beyond the capabilities of current models \citep{mallen_subversion_2024}, it may become a threat in future. As such,  
future research should investigate whether certain types of manipulation become significantly more effective when deployed by multiple coordinated models and whether control protocols remain effective against such coordinated efforts.